\documentclass[10pt,twocolumn,letterpaper]{article}

\pdfoutput=1

\usepackage{cvpr}
\usepackage{times}
\usepackage{epsfig}
\usepackage{graphicx}
\usepackage{caption}
\usepackage{subcaption}
\usepackage{amsmath}
\usepackage{amssymb}

\usepackage[pagebackref=true,breaklinks=true,letterpaper=true,colorlinks,bookmarks=false]{hyperref}

\cvprfinalcopy 

\def\cvprPaperID{4894} 

\ifcvprfinal\fi
\begin{document}

\title{Street Scene: A new dataset and evaluation protocol for video anomaly detection}

\author{Bharathkumar Ramachandra\\
North Carolina State University\\
Raleigh, NC\\
{\tt\small bramach2@ncsu.edu}
\and
Michael J. Jones\\
Mitsubishi Electric Research Labs\\
Cambridge, MA\\
{\tt\small mjones@merl.com}
}

\maketitle

\begin{abstract}
   Progress in video anomaly detection research is currently slowed by
   small datasets that lack a wide variety of activities as well as
   flawed evaluation criteria.  This paper aims to help move this
   research effort forward by introducing a large and varied new
   dataset called Street Scene, as well as two new evaluation criteria
   that provide a better estimate of how an algorithm will perform in
   practice.  In addition to the new dataset and evaluation criteria,
   we present two variations of a novel baseline video anomaly
   detection algorithm and show they are much more accurate on Street
   Scene than two well known algorithms from the literature.
\end{abstract}


\section{Introduction}

Surveillance cameras are ubiquitous, and having humans monitor them
constantly is not practical.  In most cases, almost all of the video
from a surveillance camera is unimportant and only unusual video
segments are of interest.  This is the main motivation for
developing video anomaly detection algorithms - to automatically find
parts of a video that are unusual and flag those for human inspection.

The problem of video anomaly detection can be formulated as follows.
Given one or more training videos from a single scene containing only
normal (non-anomalous) events, detect anomalous events in testing
video from the same scene.  Providing training video of normal
activity is necessary to define what is normal for a particular scene.
By {\em anomalous event}, we mean a spatially and temporally localized
segment of video that is significantly different from anything
occurring in the training video.  What exactly is meant by
``significantly different'' is difficult to specify and really depends
on the target application.  This difference could be caused by several
factors, most commonly unusual appearance or motion of objects in the
video.

It is important to point out that while many papers formulate the
video anomaly detection problem consistently with our description
above
(\cite{AdamEtAl2008,KimGrauman2009,BenezethEtAl2009,cong_abnormal_2013,SaligramaChen2012,LuEtAl2013,xu_learning_2015,HinamiEtAl2017,RavanbakhshEtAl2018}),
there are other papers that use different formulations
(\cite{SultaniEtAl2018,IonescuEtAl2017,AdamEtAl2008,LiuEtAl2018,HasanEtAl2016,UMN,IonescuEtAl2019}).
For example, some papers do not assume that the normal videos all come
from the same scene.  Sultani et al.~\cite{SultaniEtAl2018}
and Liu et al.~\cite{LiuEtAl2018} both use normal data coming from
many different scenes to build a single model.  Allowing multiple scenes to define
normal data restricts the types of anomalies that are possible to detect.  For instance, using multiple scenes to define normal data excludes anomalies such as a person walking in a restricted area.  The only way to learn that a particular spatial region of a scene is a restricted area is to see normal video of that particular scene and observe the absence of people walking in that area.  Video from other cameras/scenes gives no information about what activities may be anomalous in some areas, but not others, in a different scene.  A single model has no way of representing, for example, that a grassy area is only a restricted area in a certain location of one scene but in a different location of another scene (unless separate models are created for each scene, in which case this is equivalent to the single scene formulation).  This is true of many activities that are only anomalous in certain areas of a particular scene (such as jaywalking, cars or bikes going the wrong direction for a particular lane, etc).  Thus, the single scene formulation leads to a qualitatively different problem than a multiple scene formulation.  Because the single scene formulation corresponds to the most common surveillance use case, we are focused on it.

Another alternative formulation only defines anomalies temporally but not spatially \cite{SultaniEtAl2018,AdamEtAl2008,UMN}.  Our perspective is that for
scenes with a lot of activity, it is important to roughly localize
anomalies both temporally and spatially, in order to have confidence
that the algorithm is detecting anomalous frames for the right reasons,
and also because localizing anomalies is helpful to humans inspecting
the output of an anomaly detection algorithm.

\begin{figure}[t]
\begin{center}
  \includegraphics[width=0.9\linewidth]{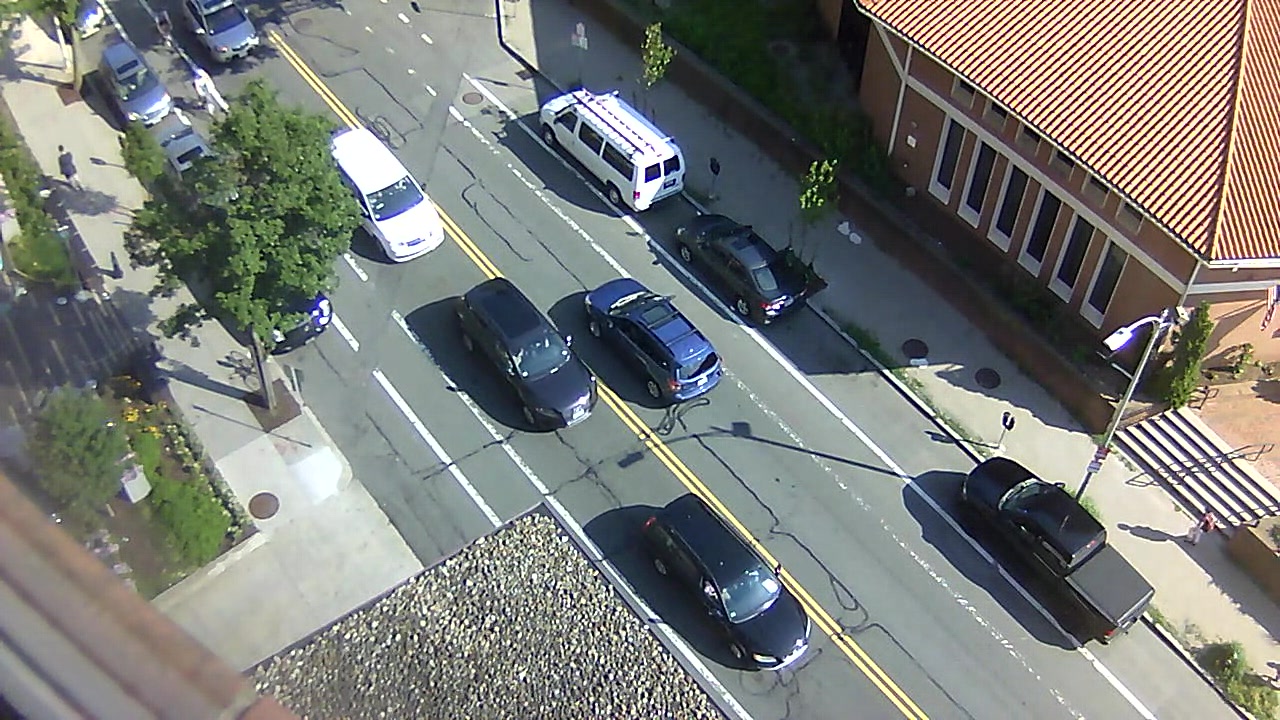}
\end{center}
\vspace{-12pt}
\caption{A normal frame from the Street Scene dataset.}
\label{fig:ex_frames}
\vspace{-9pt}
\end{figure}

After working on this problem, we think there are deficiencies in
existing datasets for the single scene formulation of video
anomaly detection.  These deficiencies include the simplicity of the
scenes for many datasets, the small number of anomalous events, the
lack of variety in anomalous events, the very low resolution of some
datasets, existence of staged anomalies in some cases, inconsistency
in annotation, and the lack of spatial ground truth (in addition to
temporal) in some cases.  Furthermore, the evaluation criteria that
have become standard practice for video anomaly detection have
problems.  Namely, the criteria do not properly evaluate
spatial localization and do not properly count false positives.  In
short, they do not give a realistic picture of how an algorithm will
perform in practice.

The goal of this paper is to shift the focus of video anomaly
detection research to more realistic datasets and more useful
evaluation criteria. We introduce a new dataset for
video anomaly detection, called Street Scene, that has more labeled
anomalous events and a greater variety of anomalies than previous
datasets for single scene anomaly detection. Street Scene contains video of a two-way urban street
including bike lanes and pedestrian sidewalks (see Figure
\ref{fig:ex_frames}).  The video is high resolution and captures a
scene with a large variety of activity.  We also suggest two new
evaluation criteria which we believe give a more accurate picture of
how video anomaly detection algorithms will perform in practice than the existing criteria.  Finally, we present two variations of a novel
algorithm which outperform two state-of-the-art algorithms on Street Scene
and set a more realistic baseline for future work to compare against.


\section{Existing Datasets and Evaluation Criteria}

There are a handful of publicly available datasets used to evaluate
video anomaly detection algorithms.  We discuss each of these below
and summarize them in Table \ref{tab:rel-datasets}.

\begin{table*}[t]
\begin{center}
  \vspace{-0.7pt}
  \resizebox{0.8\linewidth}{!}{
\begin{tabular}{|c|r|r|r|r|r|r|r|} \hline
  {\bf Dataset} & {\bf Total} & {\bf Training} & {\bf Testing} & {\bf Anomalous} & {\bf Anomaly} & {\bf Ground} & {\bf Resolution}\\
   & {\bf Frames} & {\bf Frames} & {\bf Frames} & {\bf Events} & {\bf Types} & {\bf Truth} & \\ \hline
  UCSD Ped1 & 14,000 & 6800 & 7200 & 54 & 5 & Spatial, Temporal & 238 x 158\\ \hline
  UCSD Ped2 & 4560 & 2550 & 2010 & 23 & 5 & Spatial, Temporal & 360 x 240\\ \hline
  Subway entrance$^{*}$ & 86,535 & 18,000 & 68,535 & 66 & 5 & Temporal &  512 x 384 \\ \hline
  Subway exit$^{*}$ & 38,940 & 4,500 & 34,440 & 19 & 3 & Temporal &  512 x 384 \\ \hline
  CUHK Avenue & 30,652 & 15,328 & 15,324 & 47 & ~5 & Spatial, Temporal & 640 x 360 \\ \hline
  UMN$^{**}$ & 3,855 & N/A & N/A & 11 & 1 & Temporal & 320 x 240\\ \hline
  \textbf{Street Scene} & \textbf{203,257} & \textbf{56,847} & \textbf{146,410} & \textbf{205} & \textbf{17} & Spatial, Temporal & 1280 x 720\\ \hline
\end{tabular}
}
\end{center}
\vspace{-12pt}
\caption{Characteristics of video anomaly detection datasets for the single scene formulation.  $^*$using 15fps $^{**}$aggregates from 3 cameras.
}
\label{tab:rel-datasets}
\vspace{-10pt}
\end{table*}

{\bf UCSD Pedestrian: } The most widely used video anomaly detection
dataset is the UCSD pedestrian anomaly dataset~\cite{LiEtAl2014} which
consists of two separate datasets containing video from two different
static cameras (labeled Ped1 and Ped2), each looking at a pedestrian
walkway.  The test videos contain 5 different types of anomalies: ``bike'', ``skater'', ``cart'', ``walk across'', and
``other''.

Despite being widely used, this dataset has various deficiencies.  One
is that it is modest in size, in terms of number of frames, total
anomalies, and number of different types of anomalies.  Another is that
all of the anomalies can be detected by only analyzing a single frame
at a time.

{\bf Subway: }
The Subway dataset \cite{AdamEtAl2008} contains two long videos of a
subway entrance and exit that mainly capture people entering and
leaving through turnstiles.  It is also actually two separate
datasets.  Anomalous activities include people jumping or squeezing
around the turnstiles, walking the wrong direction, and a person
cleaning the walls.  Because only two long videos are provided, there
are various ambiguities with this dataset such as what frame rate to
extract frames, which frames to use as train/test and exactly which
frames are labeled as anomalous.  Also, there is no spatial ground truth provided.

{\bf CUHK Avenue: }
Another widely used dataset is called CUHK Avenue~\cite{LuEtAl2013}.
This dataset consists of short video clips taken from a single outdoor
surveillance camera looking at the side of a building with a pedestrian
walkway in front of it.  The main activity consists of people walking
and going into or out of the building.  Anomalies are
mostly staged and consist of actions such as a person throwing papers
or a backpack into the air, or a child skipping across the walkway.  Like UCSD, this dataset also has a small number and variety of anomalies.

{\bf UMN: }
The UMN dataset contains 11 short clips of 3 scenes of people
meandering around an outdoor field, an outdoor courtyard, or an indoor
foyer.  In each of the clips the anomaly consists of all of the people
suddenly running away, hinting at a frantic evacuation scenario.  The
scene is staged and there is one anomalous event per clip.  There is
no clear specification of a split between training and testing frames
and anomalies are only labeled temporally.

{\bf Other Datasets: }
There are two other datasets that should be mentioned although they do
not fall under the single scene formulation of video anomaly
detection.  One is the ShanghaiTech dataset introduced in a paper by
Liu et al.~\cite{LiuEtAl2018}.  It consists of 13 different scenes
each with multiple training and testing sequences. The dataset is intended to be
used to learn a single model and thus does not follow the single scene formulation.  While it is conceivable to treat it as 13
separate datasets, this is problematic since many of the videos for a
particular scene have significant changes in viewpoint and some have very little training video.  Furthermore, treating it as separate
datasets would yield an average of 10 anomalous events per scene which is very small.

Another dataset from Sultani et al.~\cite{SultaniEtAl2018} (the UCF-Crime dataset) contains a
large set of internet videos taken from hundreds of different
cameras.  This dataset is intended for a very different formulation of
video anomaly detection more akin to activity detection.  In their
formulation, both anomalous and normal video is given for training.  The dataset consists of videos from many scenes labeled with predefined
anomalous activities as well as video with only "normal" activities.
For testing, only temporal labels are available, meaning spatial
evaluation cannot be done.  While this dataset is interesting, it is
for a very different version of the problem and is not applicable to
the single scene version that we are concerned with here.

General video surveillance/recognition datasets such as \cite{Liu2017Generalized,godil2019activities,nghiem2007etiseo,oh2011large} have not been used to evaluate video anomaly detection since they are not specifically curated for this purpose and do not contain sufficient ground truth annotations.

\subsection{Evaluation Criteria}

\label{sec:old_criteria}

Almost every recent paper for video anomaly detection
\cite{mahadevan_anomaly_2010,mehran_abnormal_2009,weixin_li_anomaly_2014,kratz_anomaly_2009,sabokrou_fully_2016,sabokrou_real-time_2015,cheng_video_2015,ma_anomaly_nodate,wu_chaotic_2010,vu_energy-based_2017,xu_learning_2015,HasanEtAl2016,cong_abnormal_2013,SaligramaChen2012,LuEtAl2013,sabokrou_deep-cascade:_2017,LiuEtAl2018,antic_video_2011,antic_spatio-temporal_2015,HinamiEtAl2017,IonescuEtAl2017,LiuLiPoczos2018,RavanbakhshEtAl2017,RavanbakhshEtAl2018,SmeureanuEtAl2017,SabokrouEtAl2017,IonescuEtAl2018,IonescuEtAl2019}
has used one or both of the evaluation criteria specified in Li et
al.~\cite{LiEtAl2014} which also introduced the UCSD pedestrian
dataset.  The first criterion, referred to as the {\em frame-level}
criterion, counts a frame with any detected anomalous pixels as a
positive frame and all other frames as negative.  The frame-level
ground truth annotations are then used to determine which detected
frames are true positives and which are false positives, thus yielding
frame-level true positive and false positive rates.  This criterion
uses no spatial localization and counts a frame as a correct detection
(true positive) even if the detected anomalous pixels do not overlap
with any ground truth anomalous pixels.  Even the authors who proposed
this criterion stated that they did not think it was the best one to
use \cite{LiEtAl2014}.  We have observed that some methods that claim
state-of-the-art performance on frame-level criterion perform poor
spatial localization in practice.

The other criterion is the {\em pixel-level} criterion and tries to
take into account the spatial locations of anomalies.  Unfortunately,
it does so in a problematic way.  The pixel-level criterion still
counts true and false positive frames as opposed to true and false
positive anomalous regions.  A frame with ground truth anomalies is
counted as a true positive detection if at least 40\% of the ground
truth anomalous pixels are detected.  Other pixels detected as
anomalous that do not overlap with ground truth are ignored.  Any
frame with no ground truth anomalies is counted as a false positive
frame if at least one pixel is detected as anomalous.  Given these
rules, a simple post-processing of the anomaly score maps makes the
pixel-level criterion equivalent to the frame-level criterion.  The
post-processing is: for any frame with at least one detected anomalous
pixel, label every pixel in that frame as anomalous.  This would
guarantee a correct detection if the frame has a ground truth anomaly
(since all of the ground truth anomalous pixels are covered) and would
not further increase the false positive rate if it does not (since one
or more detected pixels on a frame with no anomalies counts as a
single false positive).  This makes it clear that the pixel-level
criterion does not reward tightness of localization or penalize
looseness of it nor does it properly count false positives since false
positive regions are not even counted for frames containing ground
truth anomalies, and a frame with no ground truth anomaly can only
have a single false positive even if an algorithm falsely detects many
different false positive regions in that frame.

Better evaluation criteria are clearly needed.

\section{Description of Street Scene}

\begin{table*}[t]
\begin{center}
  \resizebox{0.9\linewidth}{!}{
\begin{tabular}{|l|c||l|c||l|c|} \hline
  {\bf Anomaly Class} & {\bf Instances} & {\bf Anomaly Class} & {\bf Instances} & {\bf Anomaly Class} & {\bf Instances} \\ \hline
  1. Jaywalking & 61 & 7. Biker on sidewalk & 7 & 13. Skateboarder in bike lane & 2\\ \hline
  2. Biker outside lane & 42 & 8. Pedestrian reverses direction & 6 & 14. Person sitting on bench & 2\\ \hline
  3. Loitering & 36 & 9. Car u-turn & 5 & 15. Metermaid ticketing car & 1\\ \hline
  4. Dog on sidewalk & 11 & 10. Car illegally parked & 5 & 16. Car turning from parking space & 1\\ \hline
  5. Car outside lane & 10 & 11. Person opening trunk & 4 & 17. Motorcycle drives onto sidewalk & 1\\ \hline
  6. Worker in bushes & 8 & 12. Person exits car on street & 3 & &\\ \hline
\end{tabular}
}
\end{center}
\vspace{-12pt}
\caption{Meta-data of anomaly classes and number of instances of each in the Street Scene dataset.}
\label{tab:anomaly_classes}
\vspace{-9pt}
\end{table*}

To address the deficiencies of existing datasets, we introduce the
Street Scene dataset.  Street Scene consists of 46 training video
sequences and 35 testing video sequences taken from a static USB
camera looking down on a scene of a two-lane street with bike lanes
and pedestrian sidewalks.  See Figure \ref{fig:ex_frames} for a
typical frame from the dataset.  Videos were collected from the camera
at various times during two consecutive summers.  All of the videos
were taken during the daytime.  The dataset is challenging because of
the variety of activity taking place such as cars driving, turning,
stopping and parking; pedestrians walking, jogging and pushing
strollers; and bikers riding in bike lanes. In addition the videos
contain changing shadows, and moving background such as a flag and
trees blowing in the wind.  There are a total of 203,257 color video
frames (56,847 for training and 146,410 for testing) each of size 1280
x 720 pixels.  The frames were extracted from the original videos at
15 frames per second.

We wanted the dataset to contain only ``natural'' anomalies, i.e. not
staged by ``actors''.  To this end, the training sequences were chosen
to meet the following conditions:

(1) If people are present, they are walking, jogging or pushing a
stroller in one direction on a sidewalk; or they are getting into or
out of their car including walking alongside their car; or
they are stopped in front of a parking meter.

(2) If a car is present, it is legally parked; or it is driving in the
appropriate direction in a car lane; or stopped in a car lane due to
traffic; or making a legal turn across traffic; or leaving/entering a
parking spot on the side of the street.

(3) If bikers are present, they are riding in the correct direction in
a bike lane; or turning from an intersecting road into a bike lane or
from a bike lane onto an intersecting road.

These conditions for normal activity imply that the following
activities, for example, are anomalous and thus do not appear in the
training videos: Pedestrians jaywalking across the road, pedestrians loitering on the sidewalk,
pedestrians walking one direction and then turning around and walking
the opposite direction, bikers on the sidewalk, bikers outside a bike
lane (except when turning into a bike lane from the intersecting
street) cars making u-turns, cars parked illegally, cars outside a car
lane (except when turning or parked, parking or leaving a parking spot).

The 35 testing sequences have a total of 205 anomalous events
consisting of 17 different anomaly types.  A complete list of anomaly
types and the number of each in the test set is given in Table
\ref{tab:anomaly_classes}, for descriptive purposes only.

The Street Scene dataset can be downloaded from http://www.merl.com/demos/video-anomaly-detection.  Ground truth annotations are provided for each testing video in the
form of bounding boxes around each anomalous event in each frame.
Each bounding box is also labeled with a track number, meaning each
anomalous event is labeled as a track of bounding boxes.  A single
frame can have more than one anomaly labeled.


\section{New Evaluation Criteria}

As discussed in Section \ref{sec:old_criteria}, the main criteria used
by previous work to evaluate video anomaly detection accuracy have
significant problems. Sabokrou et al.~\cite{SabokrouEtAl2017} also recognized the problems with the standard criteria and proposed the {\em Dual Pixel Level} criteria. While this is an improvement, it still cannot correctly count true positives and false positives in frames with (a) multiple anomalies, (b) both true positive as well as false positive detections and (c) multiple false positive detections. A good evaluation criterion should measure the fraction of anomalies an algorithm can detect and the number of false positive regions an algorithm can be expected to mistakenly find per frame.

Our new evaluation criteria are informed by the following
considerations.  Similar to object detection criteria, using the
intersection over union (IOU) between a ground truth anomalous region
and a detected anomalous region for determining whether an anomaly is
detected is a good way to insure rough spatial localization.  For
video anomaly detection, the IOU threshold should be low to allow some
imprecision in localization because of issues like imprecise labeling
(bounding boxes) and the fact that some algorithms detect anomalies
that are close to each other as one large anomalous region which
should not be penalized.  Similarly, shadows may cause larger anomalous
regions than what are labeled.  We do not think such larger than
expected anomalous-region detections should be penalized.  We use an
IOU threshold of 0.1 in our experiments.

Also, because a single frame can have multiple ground-truth anomalous
regions, correct detections should be counted at the level of anomalous regions, not frames.

False positives should be counted for each falsely detected anomalous
region, i.e. by each detected anomalous region that does not
significantly overlap with a ground truth anomalous region.  This
allows more than one false positive per frame and also false
positives in frames with ground truth annotations, unlike the previous
criteria.

In practice, for an anomaly that occurs over many frames, it is
important to detect the anomalous region in at least some of the
frames, but it is usually not important to detect the region in every
frame in the track.  This is especially true considering the
ambiguities for when to begin and end an anomalous track mentioned
earlier, and in cases where anomalous activity is severely occluded for a few frames. Because the Street Scene dataset provides track numbers for each
anomalous region which uniquely identify the event to which an
anomalous region belongs, it is easy to compute such a criterion.  The new criteria resulting from these considerations are similar to evaluation criteria used in object detection and object tracking \cite{pascal-voc-2012,Ellis2002} but similar criteria have not been used for video anomaly detection in the past.

\subsection{Track-Based Detection Criterion}

The track-based detection criterion measures the track-based detection
rate (TBDR) versus the number of false positive regions per frame.

A ground truth track is considered detected if at least a fraction
$\alpha$ of the ground truth regions in the track are detected.

A ground truth region in a frame is considered detected if the
intersection over union (IOU) between the ground truth region and a
detected region is greater than or equal to $\beta$.
\begin{equation}
\vspace{-6pt}
\mbox{TBDR} = \frac{\mbox{num. of anomalous tracks detected}}{\mbox{total num. of anomalous tracks}}.
\end{equation}
A detected region in a frame is a false positive if the IOU between it
and every ground truth region in that frame is less than $\beta$.
\begin{equation}
\vspace{-6pt}
\mbox{FPR} = \frac{\mbox{total false positive regions}}{\mbox{total frames}}
\end{equation}
where FPR is the false-positive rate per frame.

Note that a single detected region can cover two or more different
ground truth regions so that each ground truth region is detected
(although this is rare).

In our experiments below, we use $\alpha=0.1$ and $\beta=0.1$.

\subsection{Region-Based Detection Criterion}

The region-based detection criterion measures the region-based
detection rate (RBDR) over all frames in the test set versus the
number of false positive regions per frame.

As with the track-based detection criterion, a ground truth region in
a frame is considered detected if the intersection over union (IOU)
between the ground truth region and a detected region is greater than
or equal to $\beta$.
\begin{equation}
\vspace{-6pt}
\mbox{RBDR} = \frac{\mbox{num. of anomalous regions detected}}{\mbox{total num. of anomalous regions}}.
\end{equation}
The RBDR is computed over all ground truth anomalous regions in all
frames of the test set.

The number of false positives per frame is calculated in the same way
as with the track-based detection criterion.

As with any detection criterion, there is a trade-off between
detection rate (true positive rate) and false positive rate which can
be captured in a ROC curve computed by changing the threshold on the
anomaly score that determines which regions are detected as anomalous.

When a single number is desired, we suggest summarizing the
performance with the average detection rate for false positive rates
from 0 to 1, i.e. the area under the ROC curve for false positive
rates less than or equal to 1.

\section{Baseline Algorithms}

\begin{figure}[t]
\begin{center}
  \includegraphics[width=0.85\linewidth]{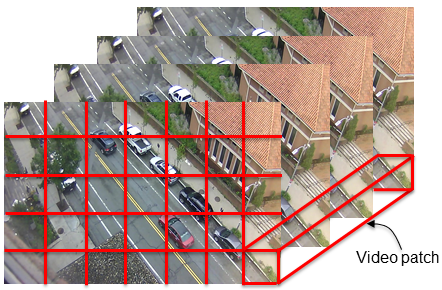}
\end{center}
\vspace{-9pt}
\caption{Illustration of a grid of regions partitioning a video frame and a video patch encompassing 4 frames.  This figure show non-overlapping regions, but in our experiments we use overlapping regions.}
\label{fig:videopatch}
\vspace{-9pt}
\end{figure}

We describe two variations of a novel algorithm for video anomaly
detection which we evaluate along with two previously published
algorithms on the Street Scene dataset in Section
\ref{sec:experiments}.  The new algorithm is based on dividing the video into spatio-temporal regions which we call video patches, storing a set of exemplars to represent the
variety of video patches occuring in each region, and then using the
distance from a testing video patch to the nearest neighbor exemplar
as the anomaly score.  As with previous work such as  \cite{IonescuEtAl2018,LuEtAl2013}, our baseline algorithm uses video patches (also called spatio-temporal cubes) as the basic building block, but differs in the features and type of model we use.

First, each video is divided into a grid of spatio-temporal regions of
size $H \times W \times T$ pixels with spatial step size $s$ and temporal step
size 1 frame.  In the experiments in Section \ref{sec:experiments} we
choose $H$=40 pixels, $W$=40 pixels, T=4 or 7 frames, and $s$ = 20
pixels.  See Figure \ref{fig:videopatch} for an illustration.

The baseline algorithm has two phases: a training or model-building
phase and a testing or anomaly detection phase.  In the model-building
phase, the training (normal) videos are used to find a set of video
patches (represented by feature vectors described later) for each
spatial region that represent the variety of activity in that spatial
region.  We call these representative video patches, {\em exemplars}.  In
the anomaly detection phase, the testing video is split into the same
regions used in training and for each testing video patch, the nearest
exemplar from its spatial region is found.  The distance to the
nearest exemplar is the anomaly score.

The only differences between the two variations are the feature vector
used to represent each video patch and the distance function used to
compare two feature vectors.

The foreground (FG) mask variation uses blurred FG masks for each
frame in a video patch.  The FG masks are computed using a background
(BG) model that is updated as the video is processed.  The BG model
used in the experiments is a very simple mean color value per pixel
although a more sophisticated model could be easily substituted.

\begin{figure}[t]
\begin{center}
  \includegraphics[width=0.8\linewidth]{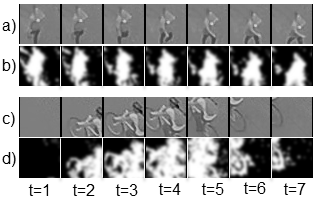}
\end{center}
\vspace{-9pt}
\caption{Example blurred FG masks, concatenated and vectorized into a feature vector.  a and c show two video patches consisting of 7 frames cropped around a spatial region.  b and d show the corresponding blurred FG masks.}
\label{fig:FG_mask}
\vspace{-15pt}
\end{figure}

The FG mask is then blurred using a Gaussian kernel to make the $L_2$
distance between FG masks more robust.  The FG mask feature vector is
formed by concatenating all of the blurred FG masks from all frames in
a video patch and then vectorizing (see Figure \ref{fig:FG_mask}).

The flow-based variation uses optical flow fields computed between
consecutive frames in place of FG masks.  The flow fields within the
region of each video patch frame are concatenated and then vectorized
to yield a feature vector twice the length of the feature vector from
the FG mask baseline (due to the dx and dy components of the flow
field).  In our experiments we use the optical flow algorithm of
Kroeger et al.~\cite{KroegerEtAl2016} to compute flow fields.

In the model building phase, a distinct set of exemplars is selected
to represent normal activity in each spatial region.  Our exemplar
selection method is straightforward.  For a particular spatial region,
the exemplar set is initialized to the empty set.  We slide a
spatial-temporal window (with step size equal to one frame) along the
temporal dimension of each training video to give a series of video
patches which we represent by either a FG-mask based feature vector or
a flow-based feature vector depending on the algorithm variation as
described above.  For each video patch, we compare it to the current
set of exemplars.  If the distance to the nearest exemplar is less
than a threshold then we discard that video patch.  Otherwise we add
it to the set of exemplars.

The distance function used to compare two exemplars depends on the
feature vector.  For blurred FG mask feature vectors, we use $L_2$
distance.  For flow-field feature vectors we use normalized $L_1$
distance:
\begin{equation}
  dist({\bf u}, {\bf v}) = \sum_i \frac{|u_i - v_i|}{|u_i| + |v_i| + \epsilon}
\end{equation}
where $u$ and $v$ are two flow-based feature vectors and $\epsilon$ is
a small positive constant used to avoid division by zero.

Given a model of normal video which consists of a different set of
exemplars for each spatial region of the video, the anomaly detection
is simply a series of nearest neighbor lookups.  For each spatial
region in a sequence of $T$ frames of a testing video, compute the
feature vector representing the video patch and then find the nearest
neighbor in that region's exemplar set.  The distance to the closest
exemplar is the anomaly score for that video patch.

This yields an anomaly score per overlapping video patch.  These are
used to create a per-pixel anomaly score matrix for each frame.  The
anomaly score for a video patch is stored in the middle frame for that
set of $T$ frames. The first $T/2 -1$ frames and the last $T/2 + 1$
frames of the testing video are not assigned any anomaly scores from
video patches and thus get all 0's.  A pixel covered by two or more
video patches is assigned the average score from all video patches
that include the pixel.

When computing ROC curves according to either of the track-based or
region-based criteria, for a given threshold, all {\em pixels} with
anomaly scores above the threshold are labeled anomalous.  Then
anomalous {\em regions} are found by computing the connected
components of anomalous pixels.  These anomalous regions are compared
to the ground truth regions according to one of the above criteria.

\section{Experiments}

\label{sec:experiments}

In addition to the two variations of our baseline video anomaly
detection method, we also tested two previously published methods.
The first is the dictionary method of Lu et al.~\cite{LuEtAl2013}
which fits a sparse combination of dictionary basis feature vectors to
a feature vector representing each spatio-temporal window of the test
video.  A dictionary of basis feature vectors is learned from the
normal training videos for each spatial region independently.  This
method reported good results on UCSD, Subway and CUHK Avenue datasets.
Code was provided by the authors.

\begin{figure*}
  \begin{center}
    \includegraphics[width=0.85\linewidth]{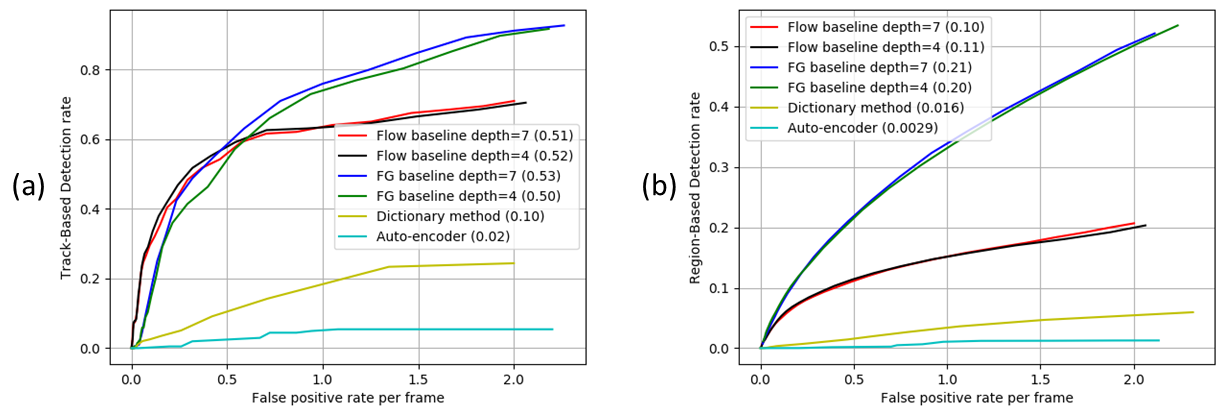}
  \end{center}    
  \vspace{-9pt}
  \caption{Track-based (a) and region-based (b) ROC curves for different methods on Street Scene}
  \label{fig:new_ROCS}
\end{figure*}

\begin{figure*}
    \begin{center}
    \includegraphics[width=0.85\linewidth]{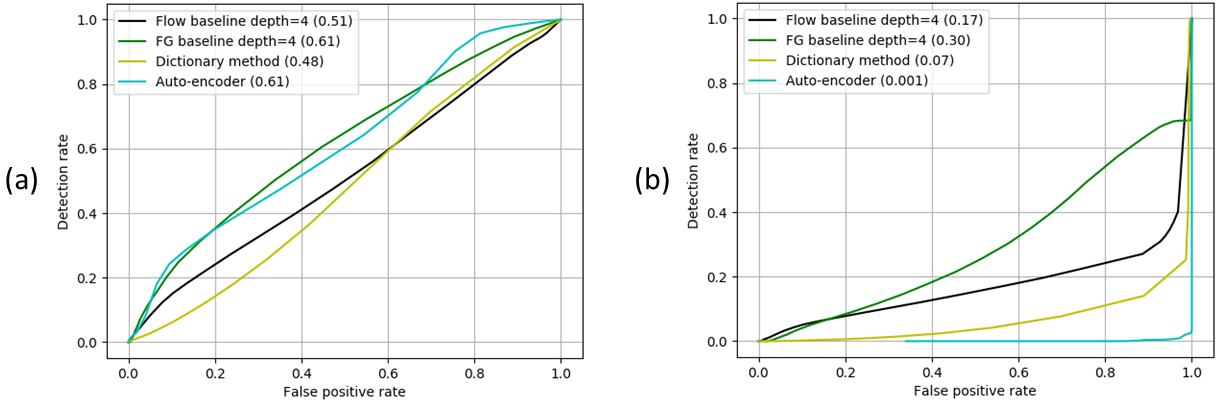}
    \end{center}
  \vspace{-9pt}
  \caption{Frame-level (a) and pixel-level (b) ROC curves for different methods on Street Scene}
  \label{fig:trad_ROCS}
  \vspace{-9pt}
\end{figure*}

The second method is from Hasan et al.~\cite{HasanEtAl2016} which uses
a deep network auto-encoder to learn a model of normal frames.  The
anomaly score for each pixel is the reconstruction error incurred by
passing a clip containing the pixel through the auto-encoder.  This
assumes that anomalous regions of a frame will not be well
reconstucted.  This method is also competitive with other
state-of-the-art results on standard datasets and evaluation criteria.
We used our own implementation of this method.

We have been unable to find code available for other algorithms, but
hope that researchers will report the results of their algorithms on
Street Scene in the near future.

Figures \ref{fig:new_ROCS} (a) and (b) show ROC curves for our
baseline methods as well as the dictionary and auto-encoder methods on
Street Scene using the newly proposed track-based and region-based
criteria.  The numbers in parentheses for each method in the figure
legends are the areas under the curve for false positive rates from 0
to 1.  Clearly, the dictionary and auto-encoder methods perform poorly
on Street Scene.  Our baseline methods do much better although there
is still much room for improvement.

While the dictionary method works well on other, smaller datasets, the
sparse dictionary model does not seem to be expressive enough to
reconstruct many normal testing video patches on the larger and more
varied Street Scene.

The auto-encoder method tries to model whole frames at once as opposed
to creating smaller models for different spatial regions.  While this
seems to work on previous datasets, it does not seem to work with the
huge variety of normal variations present in Street Scene.

Our baseline algorithms perform reasonably well on Street Scene.  They
store a large set of exemplars (typically between 1000 and 3000
exemplars) in regions where there is a lot of activity such as the
street, sidewalk and bike lane regions.  On other regions such as the
building walls or roof tops, only a single exemplar is stored.

For the two baseline variations using the track-based criteria, the
flow-based method does best for low false-positive rates (arguably the
most important part of the ROC curve).  The flow field provides more
useful information than FG masks for most of the anomalies (the main
exception being loitering anomalies which are discussed below).  The
FG-based method does better using the region-based criterion.  The
number of frames used in a video patch (4 or 7) does not have a large
effect on either variation.

The baseline algorithms do best at detecting anomalous activities such
as jaywalking, illegal u-turn, and bikers or cars outside their
lanes because these anomalies have distinctive motions compared to the
typical motions in the regions where they occur.

The loitering anomalies (and other largely static anomalies such as
illegally parked cars) are the most difficult for the baseline methods
because they do not contain any motion except at the beginning in
which a walking person transitions to loitering.  For the flow-based
method, the loitering anomalies are completely invisible.  For the
FG-based method, the beginning of the loitering anomaly is visible
since the BG model takes a few frames to absorb the motionless person.
This is the main reason why the flow-based method is worse than the
FG-based method for higher detection rates.  The FG-based method can
detect some of the loitering anomalies while the flow-based method
cannot.

A similar effect explains the region-based results in which the
FG-based method does better than the flow-based method.  The loitering
and other ``static'' anomalies make up a disproportionate fraction of
the total anomalous regions because many of them occur over many
frames.  The FG-based method detects some of these regions while the
flow-based method misses essentially all of them.  So even though the
flow-based method detects a greater fraction of all anomalous {\em
  tracks} (at low false positive rates) it detects a smaller fraction
of all anomalous {\em regions}.

\begin{figure}[t]
\begin{center}
  \includegraphics[width=1.0\linewidth]{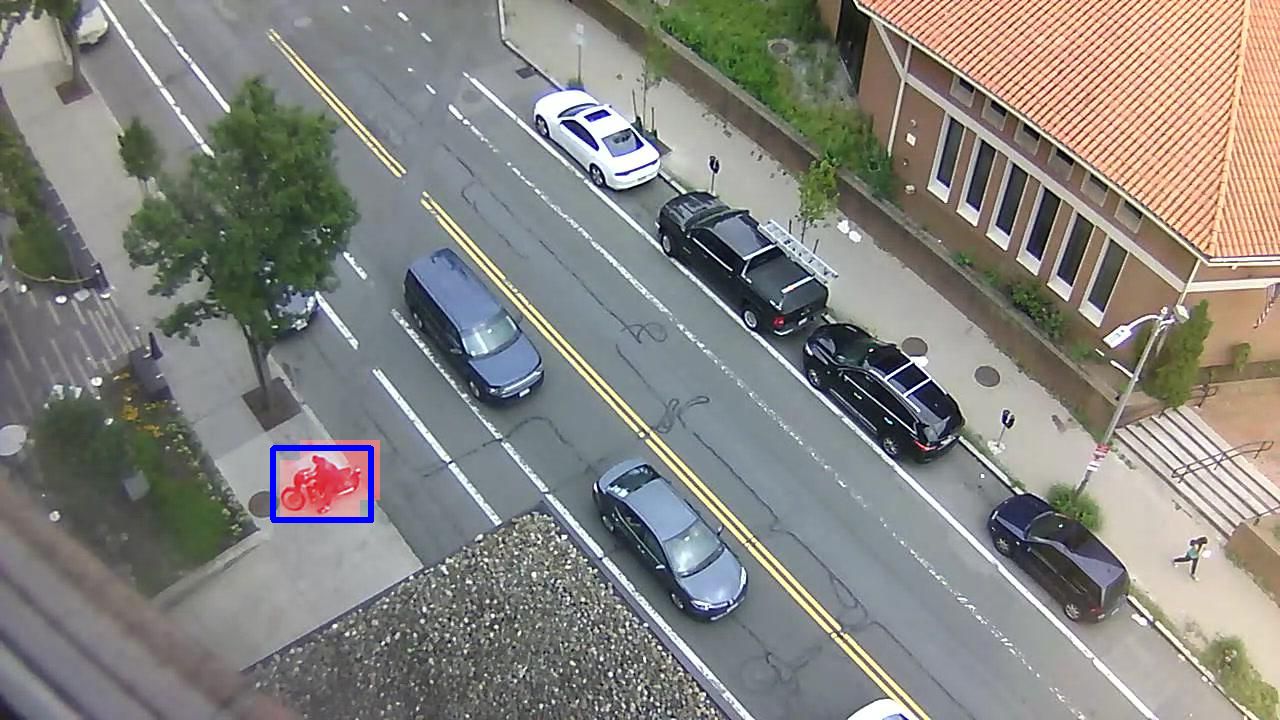}
\end{center}
\vspace{-12pt}
\caption{Detection result for flow baseline showing correctly detected motorcycle driving onto the sidewalk.}
\label{fig:motorcycle}
\vspace{-9pt}
\end{figure}

\begin{figure}[t]
\begin{center}
  \includegraphics[width=1.0\linewidth]{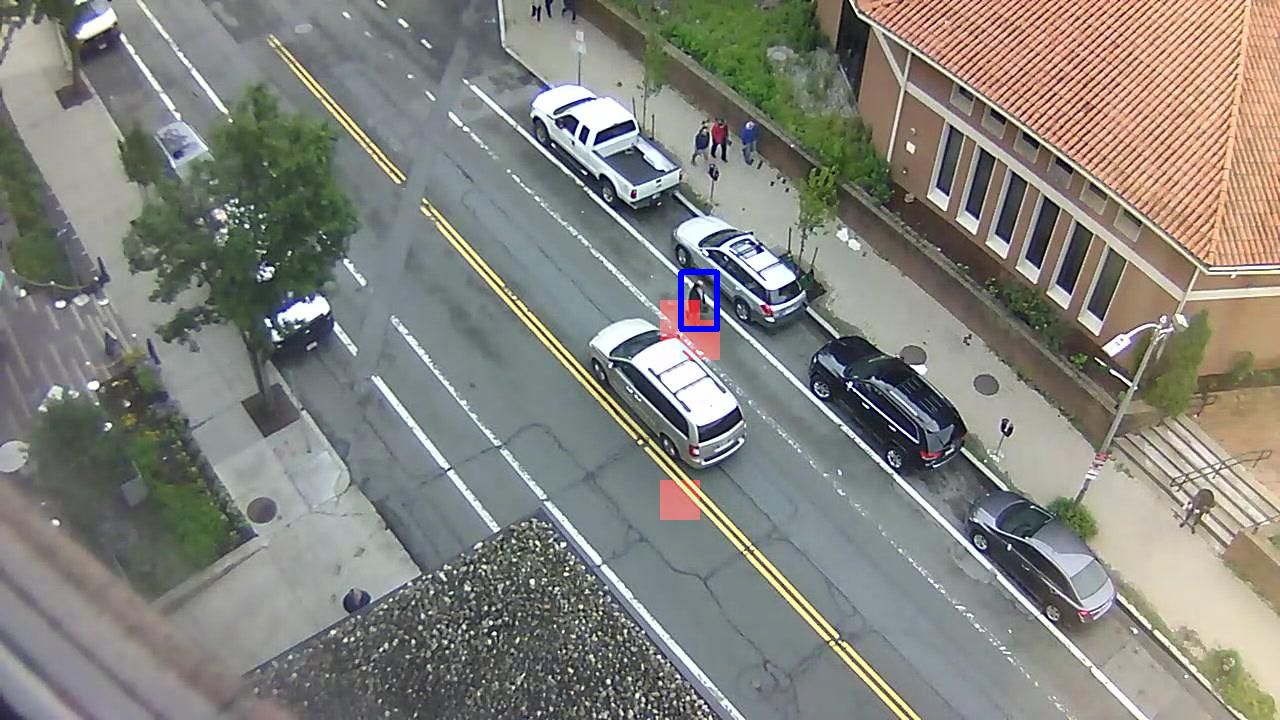}
\end{center}
\vspace{-12pt}
\caption{Detection result for flow baseline that is counted as missed detection but no false positive by pixel-level criterion and is counted as one correct detection and one false positive by the track-based and region-based criteria.}
\label{fig:jaywalkFP}
\vspace{-9pt}
\end{figure}

Some visualizations of the detection results for the flow-based method
(using T=4) are shown in Figures \ref{fig:motorcycle} and
\ref{fig:jaywalkFP}.  In the figures, red tinted pixels are anomaly
detections and blue boxes show the ground truth annotations.  Figure
\ref{fig:motorcycle} shows the correct detection of a motorcycle that
rides onto a sidewalk.  Figure \ref{fig:jaywalkFP} shows a detected
jaywalker as well as a false positive region.

\begin{figure}[t]
\begin{center}
  \includegraphics[width=1.0\linewidth]{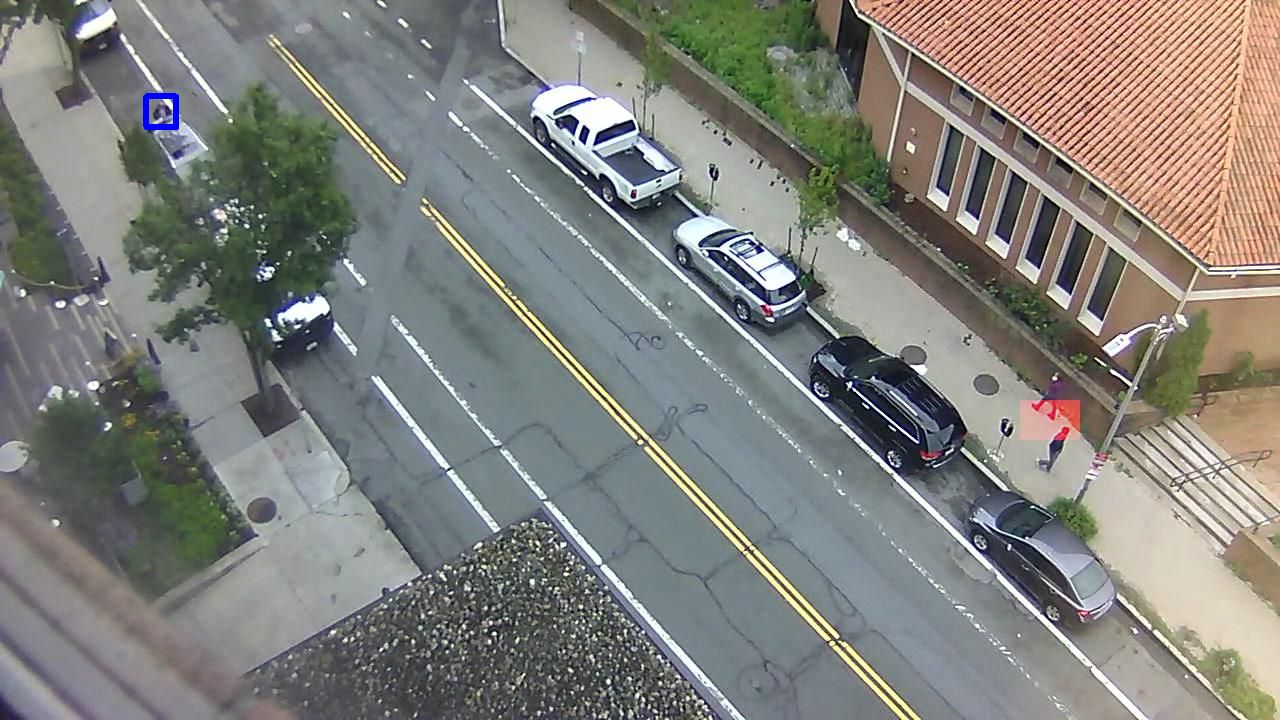}
\end{center}
\vspace{-12pt}
\caption{Detection result for flow baseline showing missed detection and false positive region that is counted as correct detection with no false positives by frame-level criterion.}
\label{fig:trunk}
\vspace{-15pt}
\end{figure}

We also show results for the two baseline algorithms as well as the
dictionary and auto-encoder methods using the traditional frame-level
and pixel-level criteria in Figures \ref{fig:trad_ROCS} (a) and (b).
We show the results for the purpose of illustrating the deficiencies
of these criteria, but not for comparison with future work.  We do not
think these criteria should be used for Street Scene going forward.
The frame-level results (which do not take spatial localization into
account) suggest that the auto-encoder method does about as well as
the foreground baseline and the dictionary method is almost as good as
the flow baseline.  However, when we look at what regions of each
frame the auto-encoder and dictionary methods actually detect as
anomalous, the accuracy is quite poor.  This can be seen in the
track-based, region-based and pixel-level ROC curves as well as by
visual inspection.  Figure \ref{fig:trunk} shows the output of the
flow baseline for a frame that contains a ``person opening trunk''
anomaly in the top, left.  The frame-level criterion counts this frame
as a correct detection even though the detected pixels are nowhere
near the ground truth anomaly but are in fact a false positive.  The
pixel-level ROC curves in Figure \ref{fig:trad_ROCS} (b) are more
reasonable and in better agreement with the track-based and
region-based ROC curves, but as mentioned earlier this criteria has
the serious flaw that a very simple post-processing of anomaly scores
would boost these curves so they are exactly the same as the
frame-level ROC curves.  Figure \ref{fig:jaywalkFP} shows an example
of a jaywalk anomaly that has fewer than 40\% of its pixels detected
and is therefore a missed detection according to the pixel-level
criterion.  This criteria also ignores a false-positive region below
the car.  The region and track-based criterion would count this as a
correct detection and one false positive.  We argue that this is a
better fit to human intuition about how this frame should be counted.

\section{Conclusions}

We have presented a new large-scale dataset and new evaluation criteria for video
anomaly detection that we hope will help to spur new innovations in
this field.  The Street Scene dataset is
a more complex scene and has almost as many anomalous events as all currently available datasets \textit{combined}. The new evaluation criteria fix the problems
with the criteria typically used in this field, and will give a more
realistic idea of how well an algorithm performs in practice.

In addition, we have presented two variations of a new video anomaly
detection algorithm as a baseline for future work to compare against;
they are straightforward and outperform two
previously published algorithms which do well on previous datasets but
not on Street Scene.  The new nearest-neighbor based algorithms may
form an interesting foundation to build on.

\textbf{Acknowledgement:} Thanks to Raju Vatsavai and Zexi Chen of NC State for help with reimplementing \cite{HasanEtAl2016}.

{\small
\bibliographystyle{ieee}
\bibliography{streetscene,Tiny}
}

\section{Supplemental Material}
\subsection{More Detection Result Visualizations}

\begin{figure}[htb]
\begin{center}
  \includegraphics[width=1.0\linewidth]{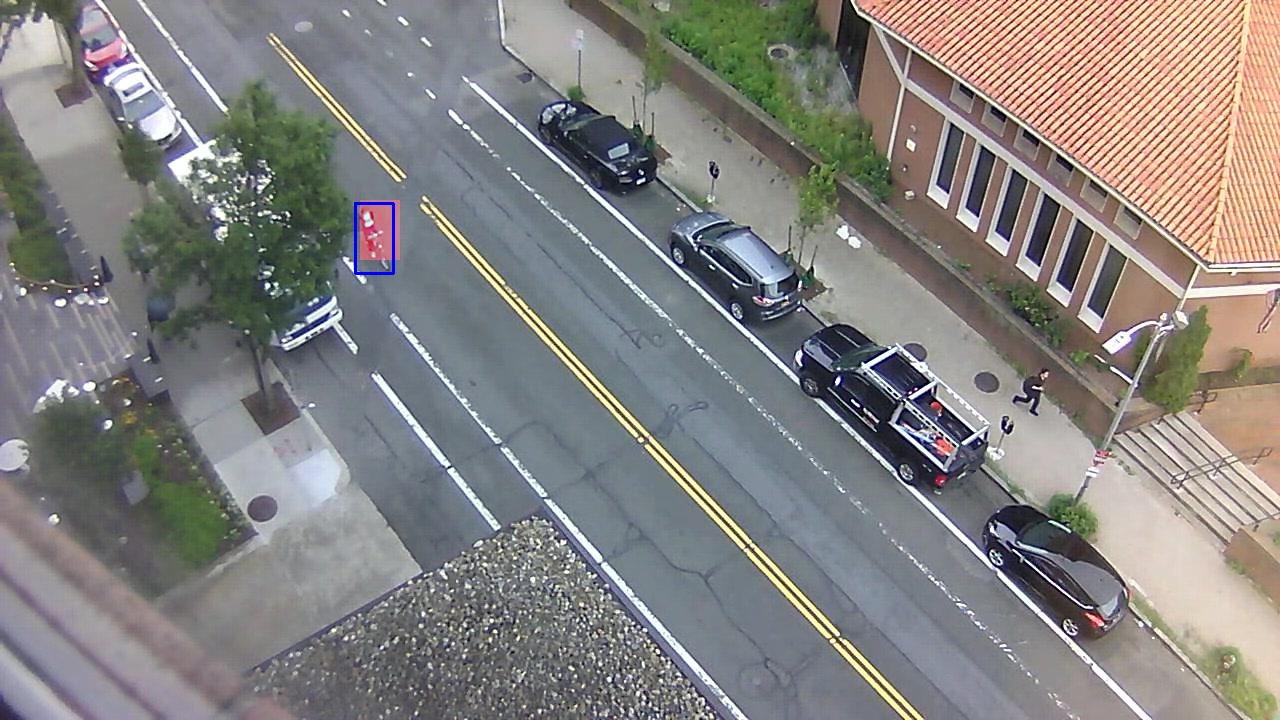}
\end{center}
\vspace{-9pt}
\caption{Detection result for flow-based method showing correctly detected biker outside of the bike lane.}
\label{fig:biker}
\vspace{-6pt}
\end{figure}

\begin{figure}[htb]
\begin{center}
  \includegraphics[width=1.0\linewidth]{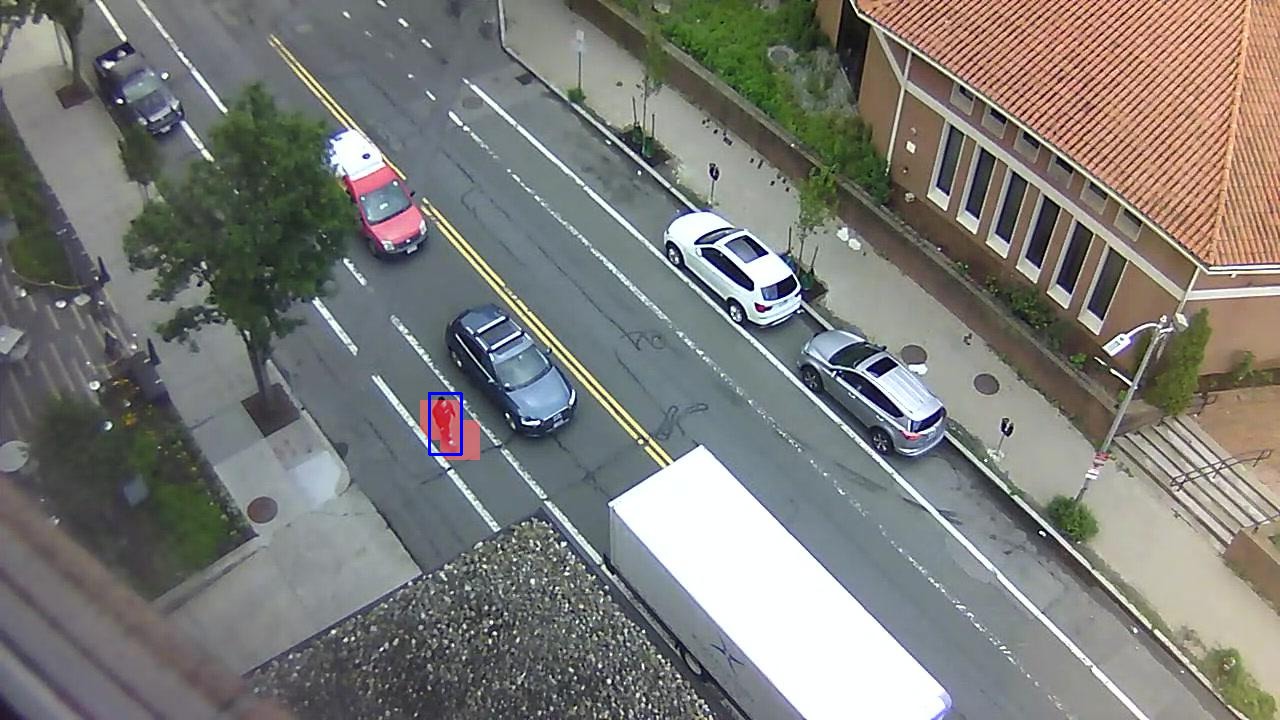}
\end{center}
\vspace{-9pt}
\caption{Detection result for flow-based method showing correctly detected skateboarder in a bike lane.}
\label{fig:skateboarder}
\vspace{-6pt}
\end{figure}

\begin{figure}[htb]
\begin{center}
  \includegraphics[width=1.0\linewidth]{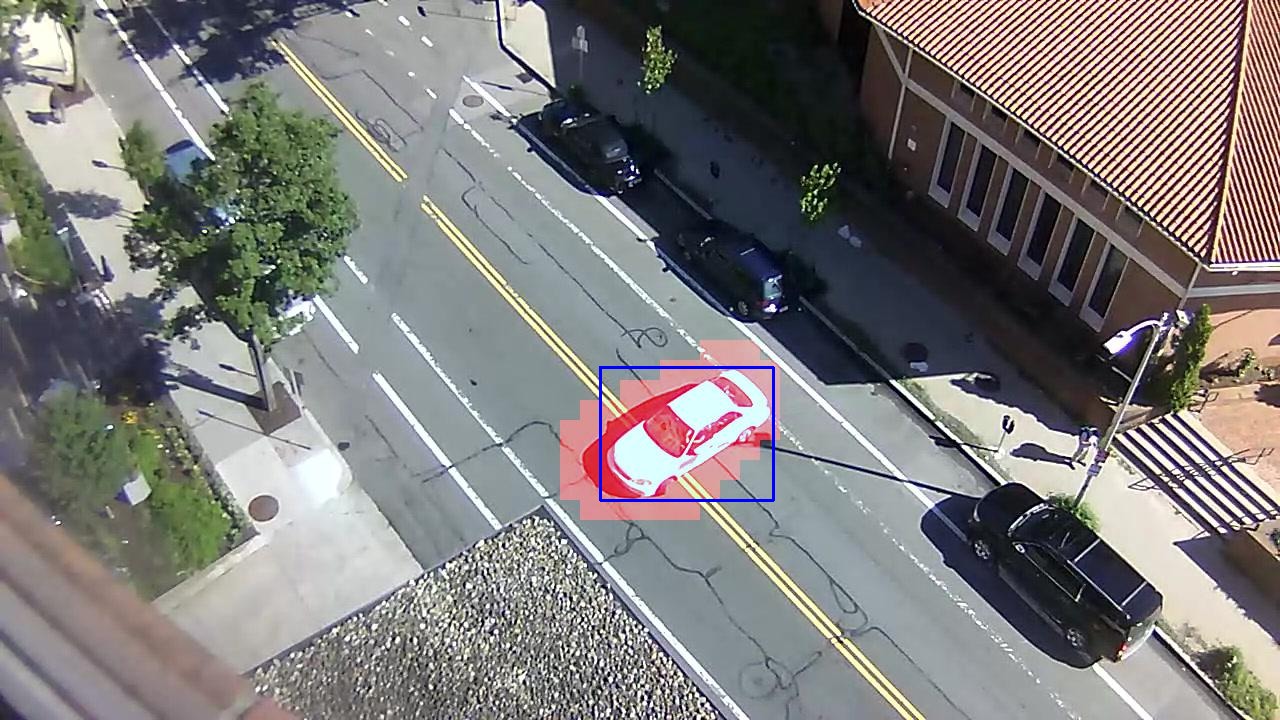}
\end{center}
\vspace{-9pt}
\caption{Detection result for flow-based method showing correctly detected car u-turn.}
\label{fig:uturn}
\vspace{-6pt}
\end{figure}

\begin{figure}[htb]
\begin{center}
  \includegraphics[width=1.0\linewidth]{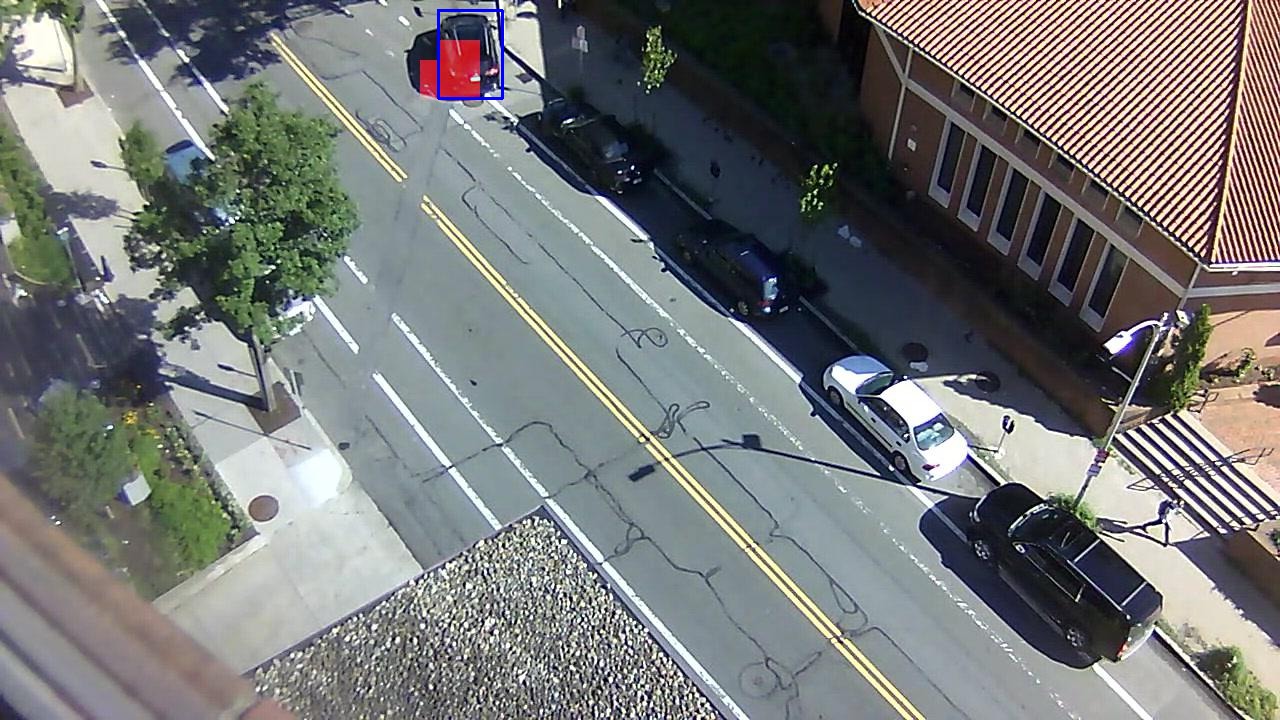}
\end{center}
\vspace{-9pt}
\caption{Detection result for flow-based method showing correctly detected illegal parking.}
\label{fig:parking}
\vspace{-6pt}
\end{figure}

\begin{figure}[htb]
\begin{center}
  \includegraphics[width=1.0\linewidth]{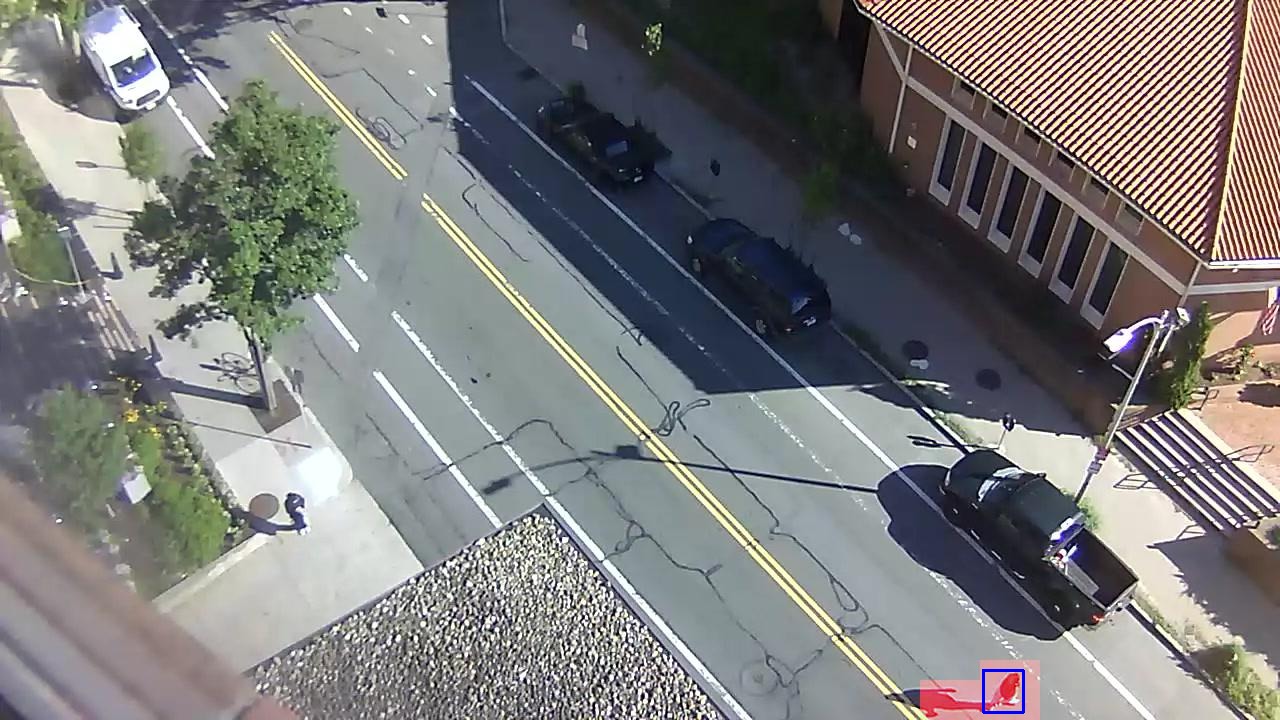}
\end{center}
\vspace{-9pt}
\caption{Detection result for flow-based method showing correctly detected jaywalker.}
\label{fig:jaywalk2}
\vspace{-6pt}
\end{figure}

\begin{figure}[htb]
\begin{center}
  \includegraphics[width=1.0\linewidth]{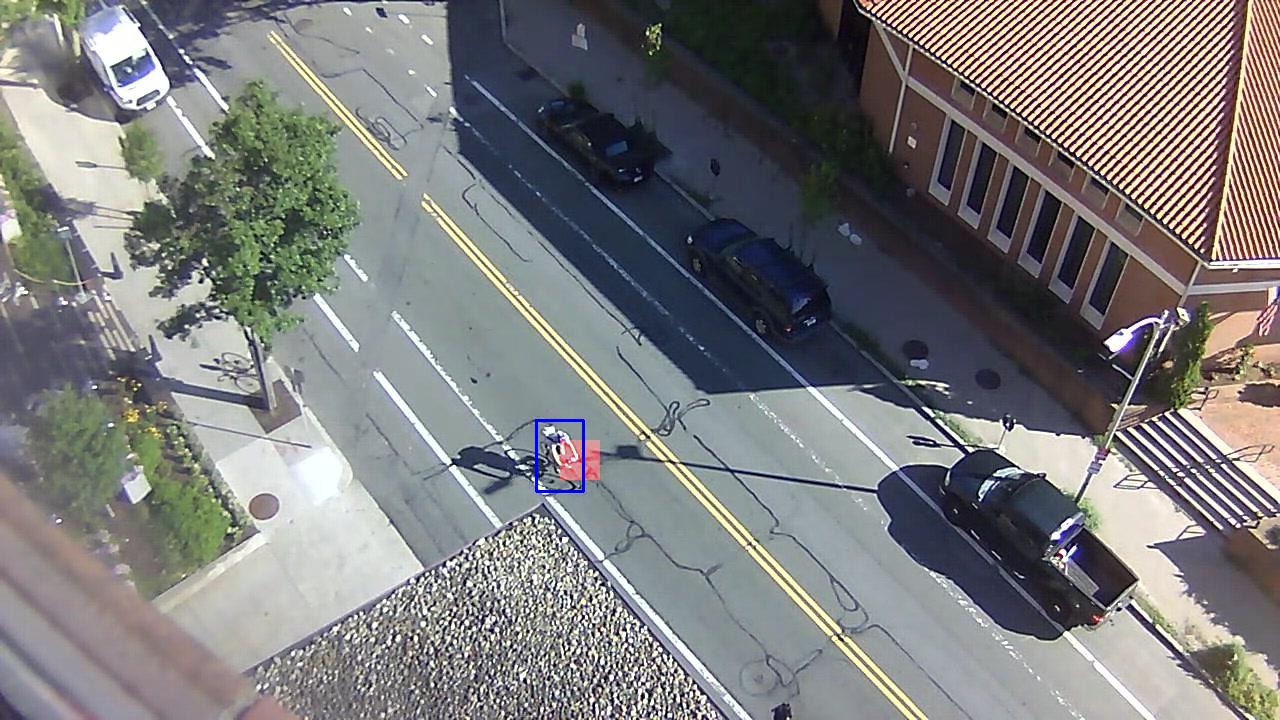}
\end{center}
\vspace{-9pt}
\caption{Detection result for flow-based method showing correctly detected biker outside of the bike lane.}
\label{fig:biker2}
\vspace{-6pt}
\end{figure}

\begin{figure}[htb]
\begin{center}
  \includegraphics[width=1.0\linewidth]{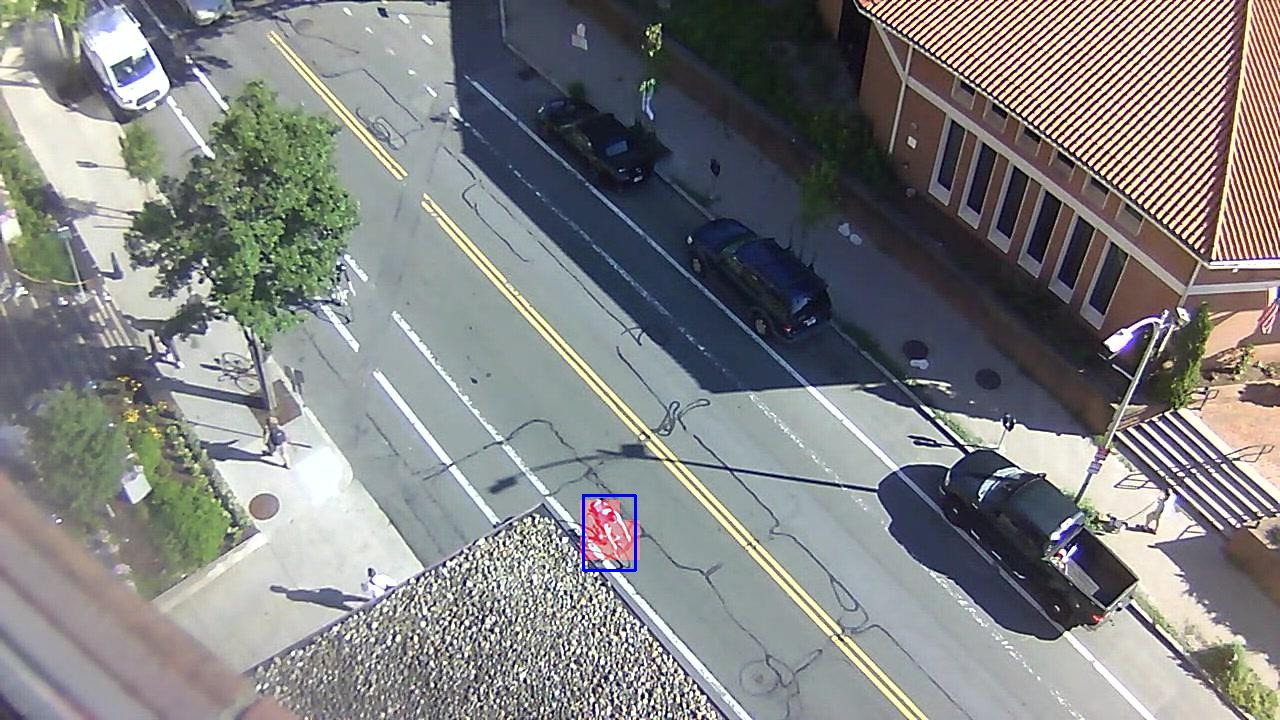}
\end{center}
\vspace{-9pt}
\caption{Detection result for flow-based method showing correctly detected biker outside of the bike lane.}
\label{fig:biker3}
\vspace{-6pt}
\end{figure}

\begin{figure}[htb]
\begin{center}
  \includegraphics[width=1.0\linewidth]{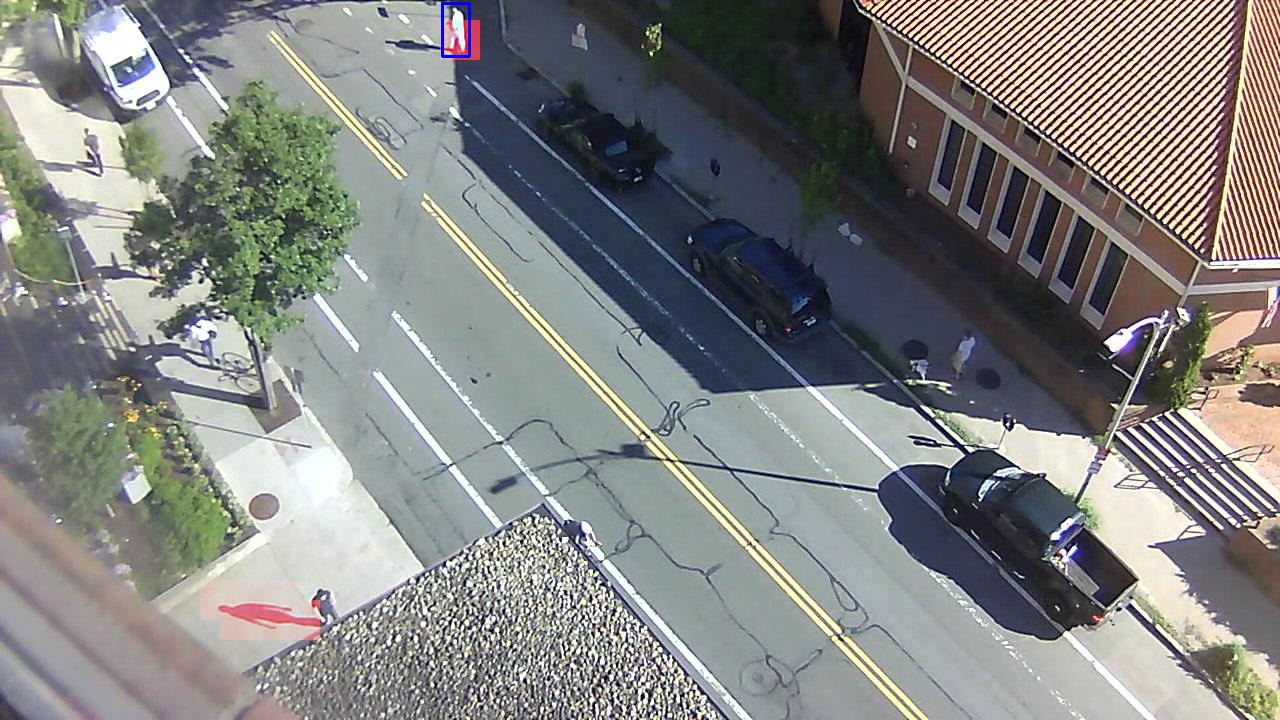}
\end{center}
\vspace{-9pt}
\caption{Detection result for flow-based method showing correctly detected jaywalker as well as a false positive.}
\label{fig:jaywalk3}
\vspace{-6pt}
\end{figure}

\begin{figure}[htb]
\begin{center}
  \includegraphics[width=1.0\linewidth]{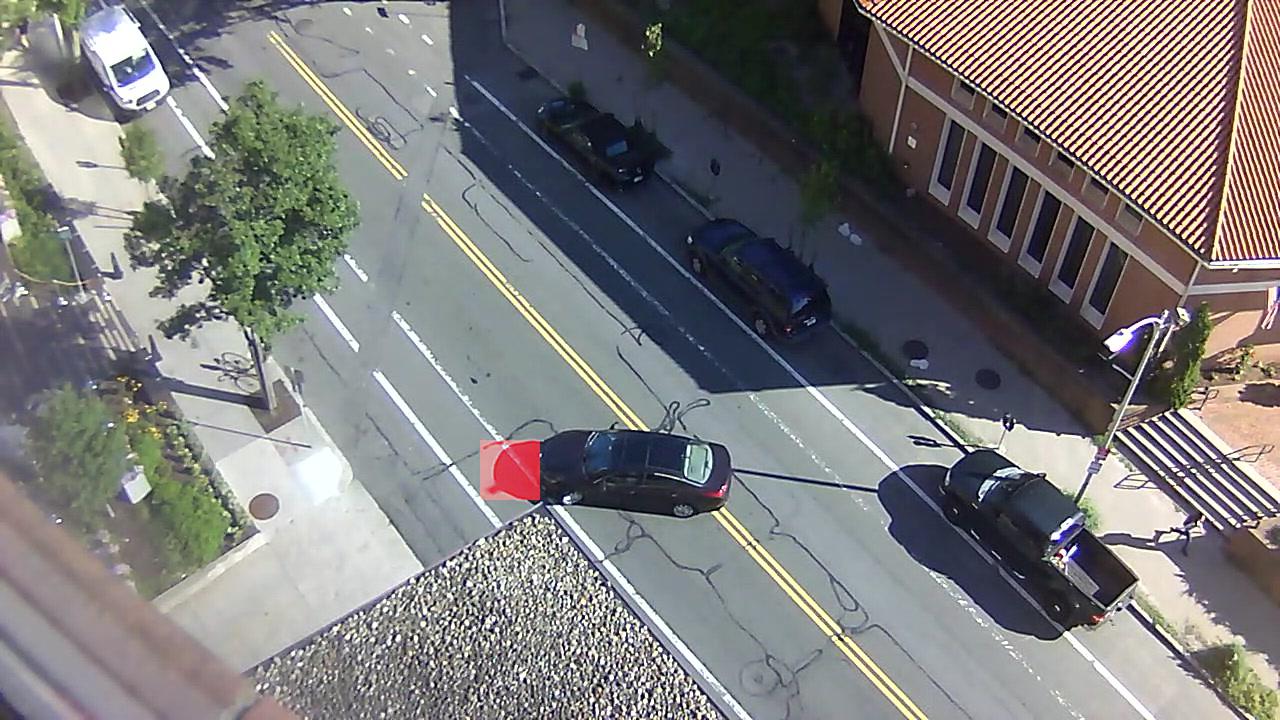}
\end{center}
\vspace{-9pt}
\caption{Detection result for flow-based method showing a false positive caused by a shadow of a car.}
\label{fig:falsepos}
\vspace{-6pt}
\end{figure}

We show more examples of our detection results using our flow-based
algorithm with $T=4$ frames in Figures \ref{fig:biker} through
\ref{fig:falsepos}.  In each of the frames shown, the red tinted
pixels are detected as anomalous by our algorithm.  The blue
rectangles show the ground truth annotations in each frame.

\subsection{Results of baseline algorithms on UCSD Ped1 and Ped2}

The baseline video anomaly detection method described in the paper is
not the focus of this paper and is not claimed to be superior to the
current state of the art on existing datasets.  The purpose is to
provide reasonable baseline results on Street Scene for future work to
compare against since the available implementations of previous
algorithms do not perform well on Street Scene.  However, readers may
be interested in how our exemplar-based algorithm performs on existing
datasets.  Table \ref{tab:ucsd} shows results for our foreground
baseline algorithm on UCSD Ped1 and Ped2 datasets using the
traditional frame-level and pixel-level criteria.  We also show
results from other recent papers for comparison.  Our results are
comparable to many recent results especially using the pixel-level
criterion.

\begin{table*}[t]
\begin{center}
    \begin{tabular}{|l|l|l|l|l|l|l|l|l|} \hline
      {\bf Method} & \multicolumn{2}{c|}{\bf Ped1 Frame-level} & \multicolumn{2}{c|}{\bf Ped1 Pixel-level} & \multicolumn{2}{c|}{\bf Ped2 Frame-level} & \multicolumn{2}{c|}{\bf Ped2 Pixel-level} \\
      & {\bf AUC} & {\bf EER} & {\bf AUC} & {\bf EER} & {\bf AUC} & {\bf EER} & {\bf AUC} & {\bf EER}\\ \hline
      Dictionary method \cite{LuEtAl2013} & 91.8\% & 15\% & 63.8\% & 43\% & - & - & - & - \\ \hline
      Autoencoder \cite{HasanEtAl2016} & 81.0\% & 27.9\% & - & - & 90.0\% & 21.7\% & - & - \\ \hline
      AMDN \cite{xu_learning_2015} & 92.1\% & 16\% & 67.2\% & 40.1\% & 90.8\% & 17\% & - & - \\ \hline
      MDT \cite{LiEtAl2014} & 81.8\% & 25\% & 44.1\% & 58.0\% & 85.0\% & 25\% & 44.0\% & - \\ \hline
      Video parsing \cite{antic_video_2011} & 91.0\% & 18\% & 83.6\% & 23\% & 92.0\% & 14\% & - & - \\ \hline
      ST Video parsing \cite{antic_spatio-temporal_2015} & 93.9\% & 12.9\% & 84.2\% & 20.5\% & 94.6\% & 10.6\% & 81.1\% & 11.2 \\ \hline
      Plug and play CNN \cite{RavanbakhshEtAl2018} & 95.7\% & 8\% & 64.5\% & -\% & 88.4\% & 18\% & - & - \\ \hline
      
      Our FG Baseline & 77.3\% & 25.9\% & 69.3\% & 39.4\% & 88.3\% & 18.9\% & 83.9\% & 23.5\% \\ \hline
    \end{tabular}
\end{center}
\vspace{-4pt}
\caption{Traditional frame-level and pixel-level results on UCSD Ped1 and Ped2.}
\label{tab:ucsd}
\vspace{-8pt}
\end{table*}

\section{Detailed List of Anomalies in Street Scene}

Tables \ref{tab:anomaly_list1} and \ref{tab:anomaly_list2} list every
annotated anomaly in the Street Scene dataset for all 35 testing
videos.  This list will be included with the dataset when it is
publicly released (along with the ground truth bounding boxes for all
frames).  The lists give a good sense of what is contained in the data
set.  It is purely for informative purposes.  The anomaly types are
not used in the evaluation criteria.

\begin{table*}[t]
\begin{center}
  \resizebox{1.0\linewidth}{!}{
    \begin{tabular}{|l|l|l||l|l|l||l|l|l|} \hline
  {\bf Test} & {\bf Anomaly} & {\bf Anomaly Type} & {\bf Test} & {\bf Anomaly} & {\bf Anomaly Type} & {\bf Test} & {\bf Anomaly} & {\bf Anomaly Type}\\
  {\bf video} & {\bf Index} & & {\bf video} & {\bf Index} & & {\bf video} & {\bf Index} & \\ \hline
  Test001 & 1 & Jaywalk & Test009 & 1 & Biker outside lane & Test015 & 4 & Biker outside lane\\ \hline
  & 2 & Worker in bushes & & 2 & Biker outside lane & & 5 & Jaywalk\\ \hline
  Test002 & 1 & Person opening trunk & & 3 & Biker on sidewalk & & 6 & Biker outside lane\\ \hline
  & 2 & Loitering & Test010 & 1 & Car u-turn & & 7 & Biker outside lane\\ \hline
  & 3 & Loitering & & 2 & Car Illegally parked & & 8 & Biker outside lane\\ \hline
  & 4 & Loitering & & 3 & Jaywalk & & 9 & Biker outside lane\\ \hline
  & 5 & Jaywalk & & 4 & Biker outside lane & Test016 & 1 & Jaywalk\\ \hline
  & 6 & Jaywalk & & 5 & Jaywalk & & 2 & Biker outside lane\\ \hline
  & 7 & Jaywalk & & 6 & Jaywalk & & 3 & Biker outside lane\\ \hline
  Test003 & 1 & Jaywalk & Test011 & 1 & Loitering & Test017 & 1 & Dog\\ \hline
  & 2 & Jaywalk & & 2 & Car u-turn & & 2 & Loitering\\ \hline
  & 3 & Jaywalk & & 3 & Biker outside lane & & 3 & Loitering\\ \hline
  Test004 & 1 & Car u-turn & & 4 & Biker outside lane & & 4 & Jaywalk\\ \hline
  & 2 & Jaywalk & & 5 & Biker outside lane & & 5 & Jaywalk\\ \hline
  & 3 & Car outside lane & & 6 & Jaywalk & & 6 & Jaywalk\\ \hline
  & 4 & Jaywalk & & 7 & Car illegally parked & & 7 & Pedestrian reverses direction\\ \hline
  & 5 & Jaywalk & & 8 & Jaywalk & & 8 & Loitering\\ \hline
  Test005 & 1 & Loitering & & 9 & Biker outside lane & & 9 & Jaywalk\\ \hline
  & 2 & Dog & & 10 & Jaywalk & & 10 & Loitering\\ \hline
  & 3 & Loitering & Test012 & 1 & Loitering & & 11 & Dog\\ \hline
  & 4 & Loitering & & 2 & Loitering & & 12 & Loitering\\ \hline
  & 5 & Jaywalk & & 3 & Car u-turn & & 13 & Biker outside lane\\ \hline
  & 6 & Loitering & & 4 & Biker outside lane & Test018 & 1 & Biker outside lane\\ \hline
  & 7 & Loitering & & 5 & Loitering & & 2 & Biker outside lane\\ \hline
  & 8 & Jaywalk & & 6 & Dog & & 3 & Biker outside lane\\ \hline
  & 9 & Jaywalk & Test013 & 1 & Dog & & 4 & Pedestrian reverses direction\\ \hline
  & 10 & Loitering & & 2 & Loitering & & 5 & Loitering\\ \hline
  & 11 & Loitering & & 3 & Biker on sidewalk & & 6 & Biker outside lane\\ \hline
  & 12 & Loitering & & 4 & Dog & & 7 & Biker outside lane\\ \hline
  & 13 & Loitering & & 5 & Loitering & & 8 & Loitering\\ \hline
  & 14 & Jaywalk & & 6 & Dog & & 9 & Metermaid ticketing car\\ \hline
  Test006 & 1 & Person sitting on bench & & 7 & Loitering & & 10 & Pedestrian reverses direction\\ \hline
  & 2 & Person opening trunk & & 8 & Loitering & & 11 & Loitering\\ \hline
  & 3 & Jaywalk & & 9 & Dog & & 12 & Biker on sidewalk\\ \hline
  Test007 & 1 & Jaywalk & & 10 & Loitering & Test019 & 1 & Person exits car on street \\ \hline
  & 2 & Skateboarder in bike lane & & 11 & Person opening trunk & Test020 & 1 & Jaywalk\\ \hline
  & 3 & Skateboarder in bike lane & Test014 & 1 & Jaywalk & & 2 & Jaywalk\\ \hline
  & 4 & Biker on sidewalk & Test015 & 1 & Car turning from parking space & & 3 & Jaywalk\\ \hline
  Test008 & 1 & Jaywalk & & 2 & Biker outside lane & Test021 & 1 & Jaywalk\\ \hline
  & 2 & Jaywalk & & 3 & Biker outside lane & & 2 & Biker outside lane\\ \hline
\end{tabular}
}
\end{center}
\vspace{-4pt}
\caption{List of anomalies labeled in each testing video}
\label{tab:anomaly_list1}
\vspace{-8pt}
\end{table*}

\begin{table*}[t]
\begin{center}
  \resizebox{1.0\linewidth}{!}{
    \begin{tabular}{|l|l|l||l|l|l||l|l|l|} \hline
  {\bf Test} & {\bf Anomaly} & {\bf Anomaly Type} & {\bf Test} & {\bf Anomaly} & {\bf Anomaly Type} & {\bf Test} & {\bf Anomaly} & {\bf Anomaly Type}\\
  {\bf video} & {\bf Index} & & {\bf video} & {\bf Index} & & {\bf video} & {\bf Index} & \\ \hline
      Test021 & 3 & Dog & Test025 & 4 & Biker outside lane & Test029 & 11 & Car outside lane\\ \hline
  & 4 & Biker outside lane & & 5 & Biker outside lane & Test030 & 1 & Car outside lane\\ \hline
  & 5 & Biker outside lane & & 6 & Biker outside lane & & 2 & Pedestrian reverses direction\\ \hline
  & 6 & Jaywalk & & 7 & Biker on sidewalk & & 3 & Car outside lane\\ \hline
  & 7 & Loitering & & 8 & Jaywalk & Test031 & 1 & Person sitting on bench\\ \hline
Test022 & 1 & Loitering & Test026 & 1 & Biker outside lane & & 2 & Biker outside lane\\ \hline
  & 2 & Dog & & 2 & Biker outside lane & & 3 & Pedestrian reverses direction\\ \hline
  & 3 & Loitering & & 3 & Car outside lane & & 4 & Jaywalk\\ \hline
  & 4 & Loitering & & 4 & Car outside lane & & 5 & Motorcycle drives onto sidewalk\\ \hline
Test023 & 1 & Dog & & 5 & Loitering & Test032 & 1 & Worker in bushes\\ \hline
  & 2 & Biker outside lane & & 6 & Biker on sidewalk & & 2 & Worker in bushes\\ \hline
  & 3 & Car u-turn & Test027 & 1 & Jaywalk & & 3 & Pedestrian reverses direction\\ \hline
  & 4 & Biker on sidewalk & & 2 & Jaywalk & & 4 & Jaywalk\\ \hline
  & 5 & Jaywalk & & 3 & Jaywalk & Test033 & 1 & Worker in bushes\\ \hline
  & 6 & Biker outside lane & & 4 & Jaywalk & & 2 & Jaywalk\\ \hline
  & 7 & Biker outside lane & Test028 & 1 & Jaywalk & & 3 & Jaywalk\\ \hline
  & 8 & Biker outside lane & & 2 & Biker outside lane & Test034 & 1 & Worker in bushes\\ \hline
  & 9 & Biker outside lane & & 3 & Car outside lane & & 2 & Jaywalk\\ \hline
  & 10 & Biker outside lane & Test029 & 1 & Illegal parking & & 3 & Jaywalk\\ \hline
Test024 & 1 & Jaywalk & & 2 & Jaywalk & & 4 & Worker in bushes\\ \hline
  & 2 & Jaywalk & & 3 & Jaywalk & & 5 & Worker in bushes\\ \hline
  & 3 & Jaywalk & & 4 & Car illegally parked & & 6 & Worker in bushes\\ \hline
  & 4 & Jaywalk & & 5 & Car outside lane & Test035 & 1 & Jaywalk\\ \hline
  & 5 & Jaywalk & & 6 & Car outside lane & & 2 & Jaywalk\\ \hline
  & 5 & Biker outside lane & & 7 & Car outside lane & & 3 & Loitering\\ \hline
  & 6 & Biker outside lane & & 8 & Person exits car on street & & 4 & Jaywalk\\ \hline
Test025 & 1 & Biker on sidewalk & & 9 & Person exits car on street & & 5 & Loitering\\ \hline
  & 2 & Jaywalk & & 10 & Person opening trunk & & 6 & Pedestrian reverses direction\\ \hline
  & 3 & Jaywalk & & & & & &\\ \hline
\end{tabular}
}
\end{center}
\vspace{-4pt}
\caption{Continued list of anomalies labeled in each testing video}
\label{tab:anomaly_list2}
\vspace{-8pt}
\end{table*}

\end{document}